\documentclass{article}

\usepackage{arxiv}

\usepackage[utf8]{inputenc} % allow utf-8 input
\usepackage[T1]{fontenc}    % use 8-bit T1 fonts
\usepackage[hidelinks]{hyperref}       % hyperlinks
\usepackage{url}            % simple URL typesetting
\usepackage{booktabs}       % professional-quality tables
\usepackage{amsfonts}       % blackboard math symbols
\usepackage{nicefrac}       % compact symbols for 1/2, etc.
\usepackage{microtype}      % microtypography
\usepackage{lipsum}		% Can be removed after putting your text content
\usepackage{graphicx}
\usepackage{natbib}
\usepackage{doi}
\usepackage{booktabs}
\usepackage{authblk}
\usepackage{subcaption}

\title{Explainable Image Quality Assessments in Teledermatological Photography}

\author[1,2]{Raluca Jalaboi}
\author[1,3,4]{Ole Winther}
\author[2]{Alfiia Galimzianova}

\affil[1]{\footnotesize Department of Applied Mathematics and Computer Science at the Technical University of Denmark, Richard Petersens Plads, Building 324, DK-2800 Kongens Lyngby, Denmark}
\affil[2]{\footnotesize Medable A/S, Silkegade 8 st, DK-1113 Copenhagen C, Denmark}
\affil[3]{\footnotesize Bioinformatics Centre, Department of Biology, University of Copenhagen, Copenhagen, Denmark}
\affil[4]{\footnotesize Center for Genomic Medicine, Rigshospitalet, Copenhagen University Hospital, Copenhagen, Denmark}

\date{}

\renewcommand{\shorttitle}{\textit{arXiv} Template}

\hypersetup{
pdftitle={Explainable Image Quality Assessments in Teledermatological Photography},
pdfauthor={Raluca~Jalaboi, Ole~Winther, Alfiia~Galimzianova},
pdfkeywords={dermatology, remote consultation, deep learning, data accuracy},
}

\begin{document}
\maketitle

\begin{abstract}
Image quality is a crucial factor in the effectiveness and efficiency of teledermatological consultations. 
However, up to 50\% of images sent by patients have quality issues, thus increasing the time to diagnosis and treatment. 
An automated, easily deployable, explainable method for assessing image quality is necessary to improve the current teledermatological consultation flow.
We introduce ImageQX, a convolutional neural network for image quality assessment with a learning mechanism for identifying the most common poor image quality explanations: bad framing, bad lighting, blur, low resolution, and distance issues.
ImageQX was trained on 26,635 photographs and validated on 9,874 photographs, each annotated with image quality labels and poor image quality explanations by up to 12 board-certified dermatologists. 
The photographic images were taken between 2017 and 2019 using a mobile skin disease tracking application accessible worldwide.
Our method achieves expert-level performance for both image quality assessment and poor image quality explanation.
For image quality assessment, ImageQX obtains a macro F1-score of $0.73\pm0.01$, which places it within standard deviation of the pairwise inter-rater F1-score of $0.77\pm0.07$. 
For poor image quality explanations, our method obtains F1-scores of between $0.37\pm0.01$ and $0.70\pm0.01$, similar to the inter-rater pairwise F1-score of between $0.24\pm0.15$ and $0.83\pm0.06$. 
Moreover, with a size of only 15 MB, ImageQX is easily deployable on mobile devices.
With an image quality detection performance similar to that of dermatologists, incorporating ImageQX into the teledermatology flow can enable a better, faster flow for remote consultations.
\end{abstract}

% keywords can be removed

\section{Introduction}
Within the past two years, consumer-facing teledermatological consultations have become much more common due to the SARS CoV-2~(COVID-19) pandemic and associated worldwide isolation measures~\citep{yeboah2021impact}. 
Teledermatological consultations are typically done via teledermatology mobile applications that require patients to photograph their skin lesions using their mobile devices, such as smartphones and tablets, and send them to dermatologists that will then diagnose the depicted skin condition remotely.
To achieve similar quality of care to an in-person consultation, high quality images are paramount~\citep{landow2014teledermatology, haque2021teledermatology}.
However, this is rarely the case: up to 50\% of patients send images taken under poor lighting conditions, that are not centered on the lesion, or that are blurry~\citep{pasquali2020teledermatology, vodrahalli2020trueimage}.

When dealing with low quality images, two main approaches exist: image denoising and image quality detection.
Image denoising processes and reconstructs noisy images such that the noise is either reduced or entirely removed.
Many denoising  methods introduce new artifacts into the images or obfuscate characteristics critical for diagnosis~\citep{lee2018performance}. Therefore, in this paper we focus on image quality detection.
By detecting the low quality images directly on the patient's mobile device, we can instruct them to retake the picture in a way that improves the quality to an acceptable level to dermatologists.
We can thus reduce the evaluation burden on dermatologists, while at the same time reducing the time to diagnosis and treatment.

Several methods for image quality detection have been previously proposed in the literature.
\cite{kim2017deep} introduce DeepIQ, a deep neural network that can identify noisy sections in an image, and compare the resulting noise maps with human assessments.
\cite{bianco2018use} propose DeepBIQ, a convolutional neural network for identifying low quality images, and report near human-level results on smartphone photos from the LIVE In the Wild challenge dataset~\citep{ghadiyaram2014crowdsourced}.
\cite{madhusudana2022image} develop CONTRIQUE, a contrastive deep learning system for creating generalizable representations using unlabeled image quality datasets.  
One common issue for all methods is the lack of a reference standard label, which limits both their training and validation rigor.
Due to this reason, they often employ unsupervised training methods and limit validation to qualitative assessment.
% Manually assessing the image quality is a time-consuming, expensive, and subjective task, with large inter-rater variations, and noisy reference labels\cite{zhai2020perceptual}.

Within teledermatology, Vodrahalli~et~al. propose a classical machine learning image quality classifier~\citep{vodrahalli2020trueimage}. Their method provides patients with explanations for the quality assessments through automated classical computer vision methods for detecting blur, lighting, and zoom issues in an image.
However, this method has several limitations: it cannot handle cases where only the background is blurry or with poor lighting, it cannot detect lesion framing issues, and it cannot discard images containing no skin.

The lack of explainability is regarded as one of the biggest obstacles towards the adoption of automated methods in medical practice~\citep{goodman2017european, kelly2019key, topol2019high}.
Gradient-based class activation maps (Grad-CAM)~\citep{selvaraju2017grad} is the most common explainability method in medical computer vision due to its ease of use, intuitive output, and low computational requirements.
Grad-CAM create class activation maps on a given convolutional layer using the backpropagation gradients -- the higher the gradient, the more important the region is to the final classification.

\begin{figure}[t!]
   \centering
       \includegraphics[width=0.55\linewidth]{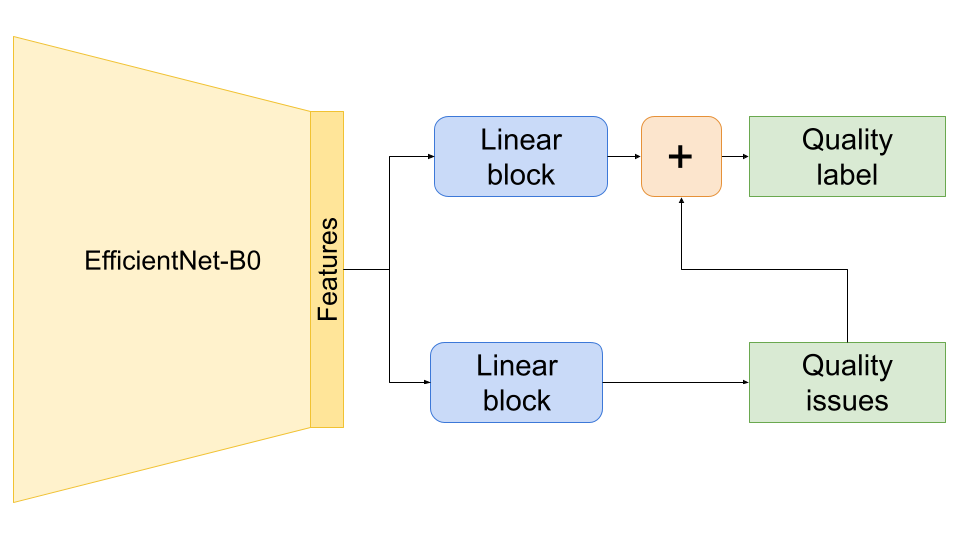}
   \caption{ImageQX network architecture. To facilitate deployment on mobile devices, we use the lightweight EfficientNet-B0 architecture as a feature extractor. A linear block, composed of a linear layer, batch normalization, and a dropout layer, is used to parse these features before predicting poor image quality explanations, i.e. \textit{bad framing, bad light, blurry, low resolution}, and \textit{too far away}. Another similar linear block parses the image features and then concatenates them with the poor image quality explanations to predict the image quality label.}
   \label{fig:architecture}
\end{figure}

In this work, we introduce ImageQX, a convolutional neural network-based method for detecting image quality.
Our novel approach uses image quality evaluations obtained from dermatologists in a teledermatology setting to learn the image quality required for a successful remote consultation.
Figure~\ref{fig:architecture} illustrates the ImageQX architecture, which learns the image quality and its explanations in an end-to-end fashion.
ImageQX was trained and validated on 36,509 images collected using a skin lesion progression tracking mobile application. 
Images were labeled by up to 12 board-certified dermatologists.
We evaluate the network performance with regard to the reference standard, and we obtain a macro F1-score of 0.73 for image quality assessment, with the per-explanation performance between 0.37 and 0.71.
ImageQX occupies only 15MB, which makes it ideal for deploying on mobile devices as a pre-filtering step during data collection.

\section{Materials and Methods}
A total of 36,509 images were collected between 2017 and 2019, using Imagine~\citep{ilab2017imagine}, a skin disease tracking mobile application available worldwide.
Self-reported user ages range between 18 and 80, and self-reported sex showing a distribution of 49\% male, 47\% female, and 4\% other.
Users span 146 countries, with images from Ukraine, United Kingdom, United States, Georgia, Russia, Albania, Kazakhstan, India, Denmark, South Africa, Bulgaria, and Israel making up 45\% of the dataset.
Images cover a wide variety of body parts.
Self-reported body part tags show that faces, arms, elbows, legs, and groin comprise the majority of images.
All patients included in this study have consented for their data to be used within a research context.

\begin{figure}[t!]
   \centering
       \includegraphics[width=0.75\linewidth]{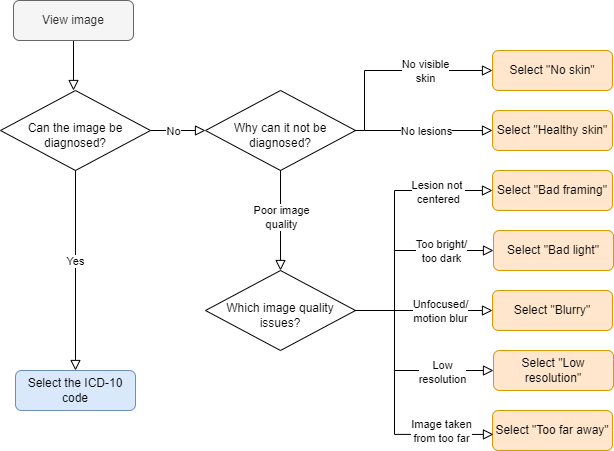}
   \caption{Labeling protocol for the ImageQX training and validation dataset. Dermatologists start by assessing whether or not the image can be diagnosed. If the image can be assessed, they diagnose it using an ICD-10 code. Otherwise, if there is no visible skin or if there are no visible lesions in the picture, the dermatologists discard the image as \textit{No skin} or \textit{Healthy skin}, respectively. Finally, if the image cannot be evaluated due to poor quality, they select one of the five investigated poor image quality explanations.}
   \label{fig:labeling_flow}
\end{figure}

\begin{figure}[t!]
    \centering
    \includegraphics[width=.95\textwidth]{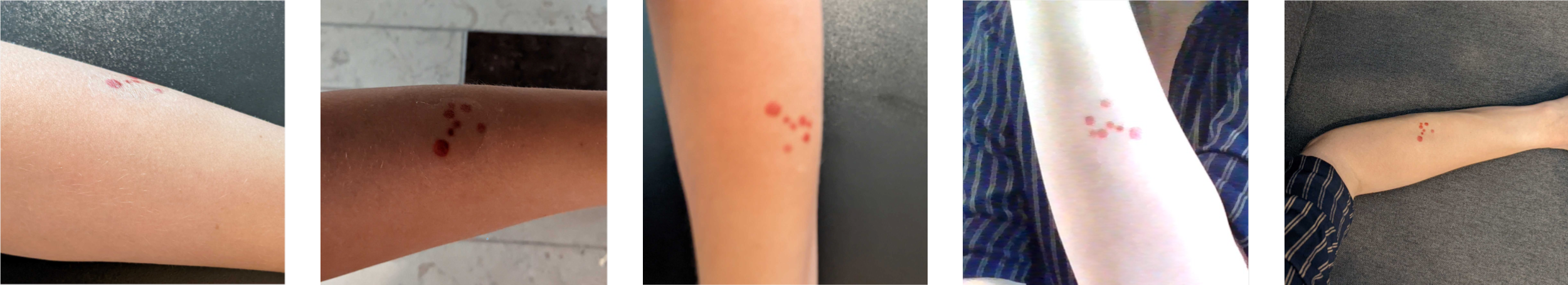}

   \caption{Illustration of poor image quality explanations that can be detected by ImageQX. (a) \emph{Bad framing}: the image was not centered on the lesion. (b) \emph{Bad light}: the lighting conditions in which the image was taken were too dark. (c) \emph{Blurry}: the image is not focused on the lesion, masking out its details. (d) \emph{Low resolution}: the image was taken with a low resolution camera and few details can be discerned. (e) \emph{Too far away}: few lesion details could be seen due to the distance from the camera. Images courtesy of the authors.}
   \label{fig:quality_issues} 
\end{figure}

Each image was evaluated by up to 12 board-certified dermatologists using an in-house labelling tool.
% (see Figure~\ref{fig:backbone}).
Dermatologists diagnosed each image with an ICD-10 code~\citep{world1992icd} whenever a lesion was present in the image and was depicted with a sufficient quality, or alternatively with one of three non-lesion labels: \emph{poor quality} when the image quality detracted from their ability to diagnose the image, \emph{healthy skin} whenever no lesions were visible, or  \emph{no skin} for images that had no dermatological relevance. 
Figure~\ref{fig:labeling_flow} outlines the protocol dermatologists followed when labeling the data, while Figure~\ref{fig:quality_issues} illustrates each poor image quality explanation included in the dataset.

We evaluate the performance of the raters and the network using sensitivity:
\begin{equation}
    Se = \frac{TP}{TP + FN},
\end{equation}
specificity:
\begin{equation}
    Sp = \frac{TN}{TN + FP},
\end{equation}
and F1-score:
\begin{equation}
    \textrm{F1-score} = \frac{2TP}{2TP+FP+FN},
\end{equation}
% with the precision $P$ is defined as
% \begin{equation}
%     P = \frac{TP}{TP + FP},
% \end{equation}
% and the recall $R$ defined as
% \begin{equation}
%     R = \frac{TP}{TP + FN},
% \end{equation}
where $TP$, $FP$, and $FN$ denote the true positives, false positives, and false negatives, respectively.
The inter-rater pairwise F1-score is calculated as the average of all dermatologist pairs, where one dermatologist is considered the reference standard while the other is considered the prediction.
For evaluating the network performance, we calculate the macro F1-score, i.e. we average the F1-scores for each class.

During training, we parse the dermatologist evaluations into four classes by merging all ICD-10 evaluations as \textit{lesion} class.
We used plurality label fusion for defining the image quality class for each image, i.e. the class selected by most dermatologists.
Alongside assessing whether the image can be evaluated or not, our proposed method also offers explanations to the \textit{poor quality} images.
To obtain the reference standard for the poor image quality explanations, we chose to mark explanations as relevant if at least one dermatologist discarded an image with that explanation.
Table~\ref{table:data_distribution} shows the distribution of labels within the dataset, while Table~\ref{table:issues_distribution} details the distribution of poor image quality explanations over the training and test sets.
Higher agreement is achieved on \textit{lesion} and \textit{no skin}, while low agreement between raters can be seen for \textit{healthy skin} and \textit{poor quality}.
Poor image quality explanations display low inter-rater agreements, with \textit{blurry} being the only one achieving an inter-rater pairwise F1-score of above 0.80.

\begin{table}[t!]
\begin{center}       
\caption{
\label{table:data_distribution}
Distribution of image quality labels over the training and test sets, including the pairwise inter-rater agreement calculated as the pairwise F1-score.} 

\begin{tabular}{lrrrr} 
\toprule
\rule[-1ex]{0pt}{3.5ex}  \textbf{Class} & \textbf{Train image count} & \textbf{Test image count} & \textbf{Pairwise train F1} & \textbf{Pairwise test F1} \\
\midrule
\rule[-1ex]{0pt}{3.5ex}  Lesion & 17,534 & 4,803 & $0.86 \pm 0.03$ & $0.84 \pm 0.08$ \\
\rule[-1ex]{0pt}{3.5ex}  No skin & 461 & 265 & $0.93 \pm 0.03$ & $0.92 \pm 0.04$ \\
\rule[-1ex]{0pt}{3.5ex}  Healthy skin & 3,903 & 2,421 & $0.62 \pm 0.10$ & $0.65 \pm 0.10$ \\
\rule[-1ex]{0pt}{3.5ex}  Poor quality & 4,737 & 2,385 & $0.63 \pm 0.08$ & $0.67 \pm 0.07$ \\
\midrule
\rule[-1ex]{0pt}{3.5ex}  Mean & 6,658.75 & 2468.5 & $0.76 \pm 0.06$ & $0.77 \pm 0.07$ \\
\bottomrule
\end{tabular}
\end{center}
\end{table}

\begin{table}[t!]
\begin{center}     
\caption{
\label{table:issues_distribution} 
Distribution of poor image quality explanations over the training and test sets, alongside the pairwise inter-rater agreement for each explanation, calculated as the pairwise F1-score.}
\begin{tabular}{lrrrr} 
\toprule
\rule[-1ex]{0pt}{3.5ex}  \textbf{Reason} & \textbf{Train image count} & \textbf{Test image count} & \textbf{Pairwise train F1} & \textbf{Pairwise train F1} \\
\midrule
\rule[-1ex]{0pt}{3.5ex}  Bad framing & 1,947 & 982 & $0.26 \pm 0.18$ & $0.24 \pm 0.15$ \\
\rule[-1ex]{0pt}{3.5ex}  Bad light & 5,144 & 2,481 & $0.63 \pm 0.07$ & $0.65 \pm 0.08$ \\
\rule[-1ex]{0pt}{3.5ex}  Blurry & 5,499 & 2,640 & $0.81 \pm 0.05$ & $0.83 \pm 0.06$ \\
\rule[-1ex]{0pt}{3.5ex}  Low resolution & 3,965 & 1,907 & $0.33 \pm 0.14$ & $0.32 \pm 0.14$ \\
\rule[-1ex]{0pt}{3.5ex}  Too far away & 936 & 497 & $0.48 \pm 0.16$ & $0.51 \pm 0.30$ \\
\midrule
\rule[-1ex]{0pt}{3.5ex}  Mean & 4,372.75 & 2,126.75 & $0.63 \pm 0.15$ & $0.64 \pm 0.18$ \\
\bottomrule
\end{tabular}
\end{center}
\end{table}

The ImageQX architecture is inspired by the DermX architecture introduced by~\cite{jalaboi2022dermx} to intrinsically learn the expert explanations, as illustrated in Figure~\ref{fig:architecture}.
EfficientNet-B0~\citep{tan2019efficientnet} was used as the feature extractor to increase the image processing speed and reduce the network size.
To increase the convergence speed, we used weights pretrained on the ImageNet dataset~\citep{deng2009imagenet}, made available by the Pytorch framework~\citep{paszke2019pytorch}.
Our network optimizes Equation 1 from~\cite{jalaboi2022dermx}:
\begin{equation}
\label{eq:dermx_loss}
    L = \lambda_D L_D + \lambda_C L_C,
\end{equation}
where $L_D$ is the categorical cross-entropy loss for the image quality label 
\begin{equation}
    L_D = - \frac{1}{ND}\sum_{i=1}^{N} \sum_{d=1}^D y_{i,d} \log \hat y_{i,d},
\end{equation}
and $L_C$ is the binary cross-entropy loss for poor image quality explanations
\begin{equation}
    L_C = -\frac{1}{NC}\sum_{i=1}^{N} \sum_{c=1}^C \left( z_{i,c} \log(\hat z_{i,c}) + (1 - z_{i,c}) \log{(1 - \hat z_{i,c})} \right).
\end{equation}
We set $\lambda_D = 1.0$ and $\lambda_C = 5.0$.
To address the imbalance in image quality labels, we used class weighted training.
Weights were set inverse proportionally to frequency in training set, as follows: 
\begin{equation}
    w_c = \min(\frac{n_{max}}{n_c}, 10.0),
\end{equation}
where $w_c$ is the weight associated with each sample in class $c$, $n_c$ is the number of samples in class $c$, and $n_{max}$ is the number of samples in the most common class.
Class weights were clipped to $10.0$ to avoid overfitting on small classes. 
This process resulted in $1.0$, $10.0$, $4.49$, and $3.70$ as weights for \textit{lesion}, \textit{no skin}, \textit{healthy skin}, and \textit{poor quality}, respectively.
The network was trained for 39 epochs with the AdamW optimizer~\citep{loshchilov2018decoupled}, cosine annealing with warm restarts~\citep{loshchilov2016sgdr}, $64$ units in each linear block, and $0.2$ dropout.
Five runs with identical hyperparameters were performed to estimate the standard deviation between training runs.

\section{Results}
Table~\ref{table:labels_performance} shows the image quality assessment performance, while Table~\ref{table:issues_performance} displays the performance on each poor image quality explanation.
The F1-scores for \textit{healthy skin} and \textit{poor quality} are within standard deviation of the inter-rater agreement, while for \textit{lesion} and \textit{no skin} the performance is slightly lower.
In the case of \textit{no skin}, this may be explained by the low amount of training data available.
For poor image quality explanations, all F1-scores except for \textit{blurry} are within standard deviation of the mean inter-rater agreement.
The high specificity visible in both image quality assessment and in poor image quality explanation suggests that deploying this network on patient phones would not negatively impact the patient experience by rejecting high quality images.
% Figure~\ref{fig:confusion_matrix} displays the image quality assessment class confusions.

\begin{table}[t!]\begin{center}       
\caption{\label{table:labels_performance} 
ImageQX performance on image quality assessment over five training runs (mean $\pm$ standard deviation). The highlighted F1-scores show the assessments where ImageQX reaches expert-level performance.} 
\begin{tabular}{lrrr} 
\toprule
\rule[-1ex]{0pt}{3.5ex}  \textbf{Class} & \textbf{Recall} & \textbf{Specificity} & \textbf{F1-score}  \\
\midrule
\rule[-1ex]{0pt}{3.5ex}  Lesion & $0.84 \pm 0.03$ & $0.78 \pm 0.04$ & \textbf{0.82 $\pm$ 0.00} \\
\rule[-1ex]{0pt}{3.5ex}  No skin & $0.76 \pm 0.05$ & $0.99 \pm 0.00$ & 0.74 $\pm$ 0.02  \\
\rule[-1ex]{0pt}{3.5ex}  Healthy skin & $0.61 \pm 0.09$ & $0.90 \pm 0.02$ & \textbf{0.63 $\pm$ 0.04} \\
\rule[-1ex]{0pt}{3.5ex}  Poor quality & $0.71 \pm 0.02$ & $0.93 \pm 0.00$ & \textbf{0.74 $\pm$ 0.01} \\
\midrule
\rule[-1ex]{0pt}{3.5ex}  Mean & $0.73 \pm 0.01$ & $0.90 \pm 0.01$ & \textbf{0.73 $\pm$ 0.01} \\
\bottomrule
\end{tabular}
\end{center}
\end{table}

\begin{table}[t!]
\begin{center}       
\caption{\label{table:issues_performance}
ImageQX performance on poor image quality explanation performance over five training runs (mean $\pm$ standard deviation). The highlighted F1-scores show the explanations where ImageQX reaches expert-level performance.} 
\begin{tabular}{lrrr} 
\toprule
\rule[-1ex]{0pt}{3.5ex}  \textbf{Reason} & \textbf{Recall} & \textbf{Specificity} & \textbf{F1-score}  \\
\midrule
\rule[-1ex]{0pt}{3.5ex}  Bad framing & $0.31 \pm 0.01$ & $0.96 \pm 0.00$ & \textbf{0.37 $\pm$ 0.01} \\
\rule[-1ex]{0pt}{3.5ex}  Bad light & $0.58 \pm 0.02$ & $0.90 \pm 0.01$ & \textbf{0.61 $\pm$ 0.00} \\
\rule[-1ex]{0pt}{3.5ex}  Blurry & $0.60 \pm 0.02$ & $0.95 \pm 0.00$ & 0.70 $\pm$ 0.01 \\
\rule[-1ex]{0pt}{3.5ex}  Low resolution & $0.47 \pm 0.02$ & $0.92 \pm 0.01$ & \textbf{0.52 $\pm$ 0.01}	 \\
\rule[-1ex]{0pt}{3.5ex}  Too far away & $0.35 \pm 0.02$ & $0.98 \pm 0.00$ & \textbf{0.42 $\pm$ 0.02} \\
\midrule
\rule[-1ex]{0pt}{3.5ex}  Mean & $0.39 \pm 0.01$ & $0.95 \pm 0.00$ & \textbf{0.45 $\pm$ 0.01} \\
\bottomrule
\end{tabular}
\end{center}

\end{table}

\begin{figure}[t!]
    \centering
    \includegraphics[width=1.\textwidth]{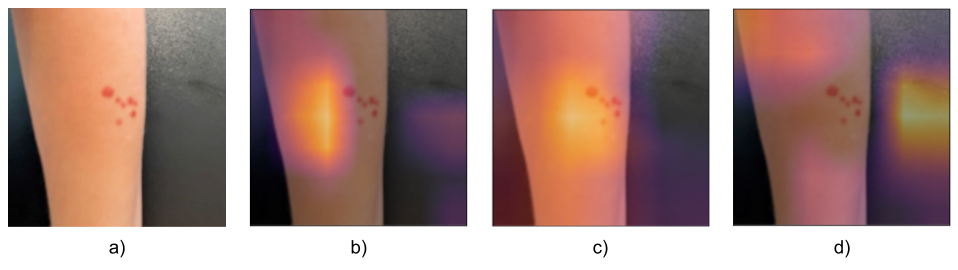}
   \caption{Grad-CAM attention maps for the blurry test image introduced in Figure~\ref{fig:quality_issues}. The image was correctly classified as poor quality. (a) the original blurry image. (b) Grad-CAM attention map for \emph{bad light}. (c) Grad-CAM attention map for \emph{blurry}. (d) Grad-CAM attention map for \emph{low resolution}. When predicting \emph{bad light}, ImageQX focuses on a slightly shaded part of the arm, while for \emph{blurry} it highlights the lesion and its surrounding area. The \emph{low resolution} prediction is based on the edges of the arm and the background. Image courtesy of the authors.}
   \label{fig:gradcams} 
\end{figure}

Figure~\ref{fig:gradcams} shows the Grad-CAM attention maps for each poor image quality explanation detected in a blurry image.
ImageQX correctly detected \textit{blurry} as one of the poor image quality explanations, focusing almost entirely on the skin area and paying more attention the lesion.
Two other explanations were also marked as present: \textit{bad light} with a focus on a slightly shaded part of the arm, and \textit{low resolution} which highlights the edges of the hand and a part of the background.

\section{Discussion}

Our data labelling process confirms the previously reported findings that poor image quality is a significant issue in teledermatology -- around 20\% of the images collected through the mobile application were labelled as poor quality by dermatologists.
However, dermatologists have low levels of agreement on which images are poor quality, with inter-rater F1-scores of $0.62 \pm 0.08$.
Explaining what makes an image poor quality is an even more difficult task, with inter-rater F1-scores varying between 0.26 and 0.81.
Similar to the dermatologists, ImageQX tends to confuse some healthy skin with skin lesions, due to common lesions (e.g. freckles, nevi) being misinterpreted as healthy skin.
Part of the disagreement can be ascribed to personal preference and level of experience with teledermatology, as some dermatologists tend to reject a larger proportion of images than others.

ImageQX reaches dermatologist-level performance on assessing the image quality on all quality assessment classes except for \textit{no skin}.
One reason for this lapse may be the low amount of training data for images with no skin.
A similar trend can be observed for poor image quality explanation, where ImageQX obtains F1-scores within a standard deviation of the inter-rater agreement for all explanations except \textit{blurry}.

Within a real world use-case, the high specificity on both the image quality assessment and poor image quality explanation suggests that the image retake burden placed on the users would be rather low -- only truly low quality or irrelevant images would be flagged for retake.
A low percentage of images with a poor quality, no skin, or healthy skin are likely to be seen by dermatologists.
Poor image quality explanations also show a high specificity, indicating that, if given proper guidance on how to fix each issue, users would find them useful in their retake attempt.
By changing the threshold for \textit{poor quality} image detection or for the image quality explanations we can further reduce the poor quality images sent to the dermatologists.
Such an intervention should be done after thorough testing with both patients and dermatologists to ensure that we identify the ideal balance between asking patients to retake the images without being too disruptive to the consultation flow.

A Grad-CAM analysis of the poor image quality explanations on a demo image (see Figure~\ref{fig:gradcams}) shows that ImageQX mostly bases its decisions on relevant areas.
The \textit{blurry} attention map is focused on the the blurry lesion, while \textit{bad light} concentrates on a slightly shaded area to the left of the lesion.
\textit{Low resolution} illustrates the debugging capabilities of Grad-CAMs: ImageQX bases its assessment primarily on the background rather than the original image.
If these attention maps were to be presented to users alongside the explanations, they could help focusing the users' attention to which sections of the image require improvement.
For example, the Grad-CAM map for \textit{blurry} suggests that the users should focus on the lesion instead of ensuring that the background is not blurred.

These findings open up several exploration avenues.
First, by adding more non-skin images from publicly available datasets we could improve the performance on the \textit{no skin} class.
This addition to the training dataset requires the data to be from the same distribution, i.e. smartphone images, to avoid in-class domain shift.
Second, to more accurately model the uncertainty inherent in the image quality assessment task, we could train ImageQX using soft labels.  
Third, we believe that by introducing a skin segmentation network as preprocessing we would avoid misclassifications due to ImageQX focusing on the background.
One drawback of this approach is the failure case of the segmentation network: if the segmentation removes the areas containing skin, the image quality assessment classifier is bound to fail.
Finally, we would like to perform a usability study to quantify the impact an on-device image quality assessment network would have on the time to diagnosis and treatment in a teledermatology setting.
Such a study would require an in-depth analysis of how to best communicate the image quality assessments and explanations to the patients.

\section{Conclusions}
Our work on ImageQX introduced several elements of novelty.
First, we quantified the dermatologist levels of agreement on what constitutes a high quality image for a teledermatological consultation and their reasoning when tagging images as low quality.
Second, we introduced ImageQX, an image quality assessor that can explain its reasons for marking an image as poor quality at an expert dermatologist level.
The added explainability component aims to facilitate the patient understanding on how to improve their images.
Moreover, with a size of only 15 MB, ImageQX can be easily packaged with a teledermatology mobile application and deployed on mobile devices, and thus incorporated as a step between users taking photos and sending them. 
Having such a network integrated in the application during the data collection step of this study would have prevented 1,819 poor quality or no skin images from being sent for assessment to the dermatologists.
In the future, we will perform a validation study to quantify the impact of introducing such a method within a consumer-facing teledermatology setting.

Our solution offers an improvement to the current consumer-facing teledermatology flow by increasing the likelihood that patients send better photos, by decreasing the time spent by dermatologists on diagnosing a single patient, and by reducing the time needed to arrive at a diagnosis and a treatment for the patients.

\bibliographystyle{unsrtnat}
\bibliography{amia}  %%% Uncomment this line and comment out the ``thebibliography'' section below to use the external .bib file (using bibtex) .

\newpage
\appendix
\section*{Appendix A}

\begin{figure}[h]
   \centering
       \includegraphics[width=0.85\linewidth]{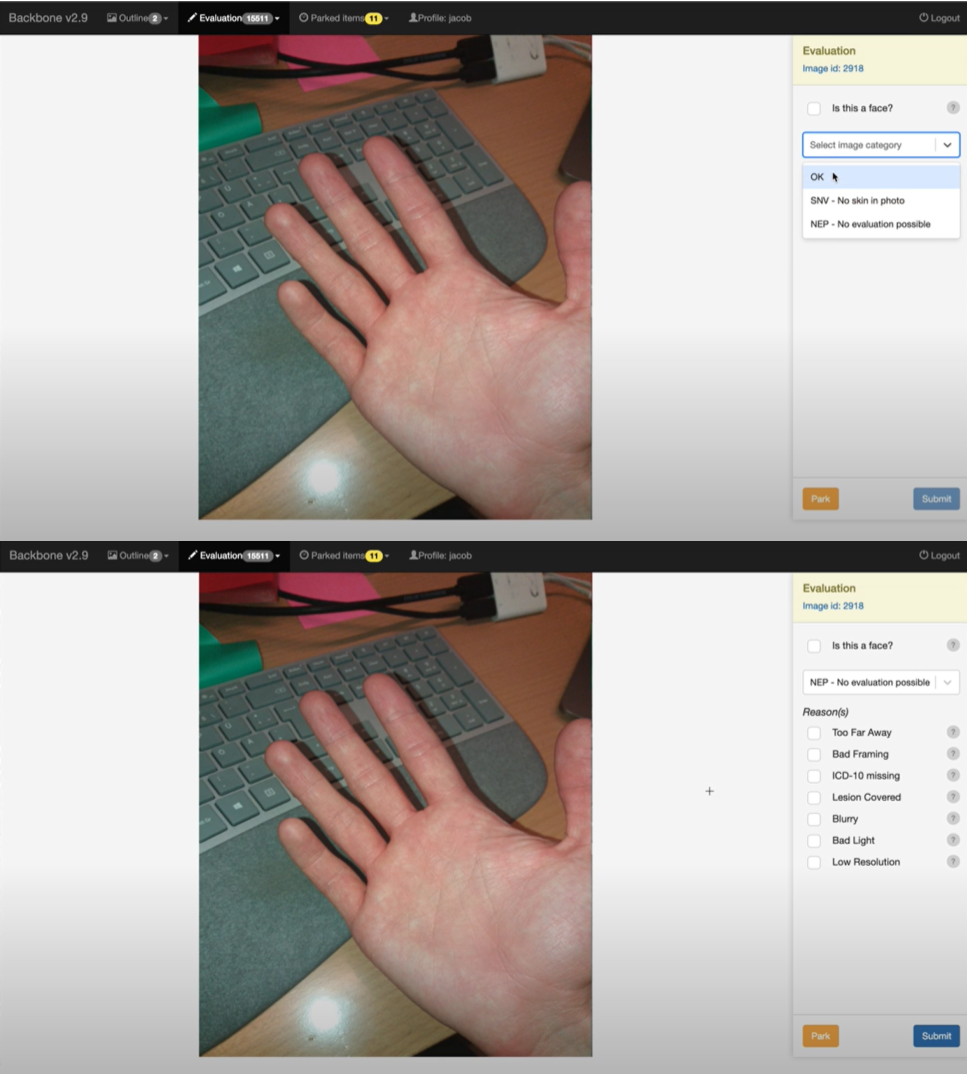}
   \caption{Two relevant screen from the labelling tool interface used by the dermatologists to evaluate the teledermatological images. 
   Dermatologists first had to select whether the image can be diagnosed (OK), if there is no skin depicted in it (SNV), or if the image is too low quality (NEP). 
   If the image was marked as low quality, they were asked to select the identified poor image quality explanations (too far away, bad framing, lesion covered, blurry, bad light, and low resolution). 
   \textit{Lesion covered} was excluded from this study due to the low number of images tagged with this explanation.
   \textit{ICD-10 missing} was used when the dermatologist could not find the ICD-10 code associated with their diagnosis.
   Image courtesy of the authors.}
   \label{fig:backbone}
\end{figure}

%%% Uncomment this section and comment out the \bibliography{references} line above to use inline references.
% \begin{thebibliography}{1}

% 	\bibitem{kour2014real}
% 	George Kour and Raid Saabne.
% 	\newblock Real-time segmentation of on-line handwritten arabic script.
% 	\newblock In {\em Frontiers in Handwriting Recognition (ICFHR), 2014 14th
% 			International Conference on}, pages 417--422. IEEE, 2014.

% 	\bibitem{kour2014fast}
% 	George Kour and Raid Saabne.
% 	\newblock Fast classification of handwritten on-line arabic characters.
% 	\newblock In {\em Soft Computing and Pattern Recognition (SoCPaR), 2014 6th
% 			International Conference of}, pages 312--318. IEEE, 2014.

% 	\bibitem{hadash2018estimate}
% 	Guy Hadash, Einat Kermany, Boaz Carmeli, Ofer Lavi, George Kour, and Alon
% 	Jacovi.
% 	\newblock Estimate and replace: A novel approach to integrating deep neural
% 	networks with existing applications.
% 	\newblock {\em arXiv preprint arXiv:1804.09028}, 2018.

% \end{thebibliography}

\end{document}